\documentclass[letterpaper]{article}
\usepackage[preprint]{aaai2027}
\usepackage[hyphens]{url}
\usepackage{graphicx}
\usepackage{natbib}
\usepackage{caption}
\usepackage{amsmath,amssymb}
\usepackage{array}
\usepackage{algorithm}
\usepackage{algorithmic}

\usepackage{newfloat}
\usepackage{listings}
\DeclareCaptionStyle{ruled}{labelfont=normalfont,labelsep=colon,strut=off} 
\floatstyle{ruled}
\newfloat{listing}{tb}{lst}{}
\floatname{listing}{Listing}

\usepackage{booktabs}

\title{Joint-Embedding Prediction of Masked Point Tubes for Self-Supervised Learning on 4D Point Cloud Videos}
\author{
Jheng-Ling Lee,
Shang-Tse Chen
}
\affiliations{
Department of Computer Science and Information Engineering, National Taiwan University\\
b11902129@csie.ntu.edu.tw, stchen@csie.ntu.edu.tw
}

\begin{document}

\maketitle

\begin{abstract}
Self-supervised representation learning for 4D point cloud videos is challenging because annotations are costly and reconstruction-based pretraining can overemphasize low-level geometric details. We propose a JEPA-style framework that learns from unlabeled spatiotemporal point clouds through latent point-tube prediction. Instead of reconstructing raw coordinates, the model masks spatiotemporal regions and predicts their target representations from visible context representations in feature space. To stabilize latent prediction, we incorporate Sketched Isotropic Gaussian Regularization, which encourages non-collapsed embeddings without relying on explicit reconstruction targets. This formulation aims to capture both spatial structure and temporal dynamics while keeping the pretraining objective aligned with downstream semantic recognition. Experiments on action and gesture recognition benchmarks show that the learned representations improve downstream fine-tuning, limited-label learning, and cross-dataset transfer. These results suggest that JEPA-style latent prediction is a promising alternative to reconstruction-centered pretraining for 4D point cloud videos.
\end{abstract}

\begin{center}
\textbf{Code}: \url{https://github.com/Leoxu3/PointTube-JEPA}
\end{center}

\section{Introduction}
\label{sec:introduction}

Self-supervised learning has become an important approach to visual representation learning because it can exploit large unlabeled corpora while reducing dependence on costly task-specific annotation. Masked autoencoding and contrastive learning have demonstrated that useful representations can be learned from images, videos, and point clouds by defining supervisory signals from the input itself~\cite{He_2022_CVPR,NEURIPS2022_e97d1081,10.1007/978-3-030-58580-8_34,10.1007/978-3-031-20086-1_35}. This property is particularly relevant for 4D point cloud videos, where each sample contains unordered 3D point sets evolving over time. Labels for action or gesture recognition are expensive to obtain, and purely supervised training can bias the representation toward the annotated downstream task instead of the full spatiotemporal structure of the signal.

Recent self-supervised methods for point cloud videos have made progress by extending masked modeling, contrastive prediction, and distillation to dynamic point sets~\cite{10378457,Shen_2023_CVPR,Zhang_2023_CVPR,zuo2025uni4dunifiedselfsupervisedlearning,nguyen2026cross4djepadensecrossmodalcorrespondence}. These approaches show that unlabeled point cloud videos contain useful cues for recognition and transfer. However, many still rely on low-level coordinate reconstruction, manually designed motion objectives, or external teacher targets. These choices may respectively bias learning toward local geometric details, introduce task-specific motion assumptions, or create dependence on the choice and domain of the teacher.

Joint-Embedding Predictive Architectures (JEPA) offer a complementary direction. Instead of reconstructing raw observations, JEPA-style methods predict target representations from context representations in latent space~\cite{10205476,bardes2024revisiting}. This formulation has been explored for images, videos, and static point clouds~\cite{10943937,hu20243djepajointembeddingpredictive}, and is well suited to 4D point cloud videos because the prediction target can focus on spatiotemporal features rather than exact coordinate recovery. A central challenge is avoiding representation collapse. While many predictive self-supervised systems use momentum target encoders, stop-gradient operations, or related stabilization heuristics, LeJEPA introduces Sketched Isotropic Gaussian Regularization (SIGReg) as an explicit embedding regularizer for stable latent prediction~\cite{balestriero2025lejepaprovablescalableselfsupervised}. This motivates a simpler formulation in which prediction and regularization are both defined directly on learned embeddings.

In this paper, we study JEPA-style self-supervised learning for 4D point cloud videos. The proposed framework tokenizes a point cloud video into local spatiotemporal point-tube tokens, masks a large subset of these tokens, and trains a predictor to infer the masked target representations from visible context representations. The objective is applied in latent space rather than in the raw coordinate space, and SIGReg regularizes the latent distribution to discourage collapse. By combining point-tube tokenization, 4D positional information, context-to-target latent prediction, and SIGReg, the method aims to learn representations that capture both spatial geometry and temporal dynamics without requiring manual motion labels or reconstruction-specific decoders.

Our contributions are summarized as follows:
\begin{itemize}
    \item We develop a teacher-free, shared-encoder JEPA formulation for 4D point cloud videos, where masked point-tube latents are predicted from visible spatiotemporal context without coordinate reconstruction.
    \item We stabilize end-to-end context-target learning with SIGReg and systematically study gradient routing, target-location conditioning, masking ratio, and regularization strength.
    \item We evaluate the learned representation under full-label fine-tuning, semi-supervised learning, few-shot recognition, transfer learning, and qualitative attention-map visualizations.
\end{itemize}

\section{Related Work}
\label{sec:related-work}

\subsection{4D Point Cloud Video Understanding}
\label{subsec:related-supervised-4d}

Supervised 4D point cloud video understanding studies how to extract spatial geometry and temporal motion from unordered point sets observed over time. Early point-based models preserve temporal structure directly in the point domain: PointLSTM formulates gesture recognition as irregular sequence modeling over neighboring points \cite{9157795}, while MeteorNet builds local spatiotemporal neighborhoods for dynamic point cloud sequences \cite{9010250}. Subsequent architectures introduce stronger 4D operators. P4Transformer combines point 4D convolution for local spatiotemporal embedding with Transformer attention over video-level features \cite{9578674}; PSTNet decomposes point spatiotemporal convolution into spatial and temporal components \cite{fan2021pstnet}; and PST-Transformer uses spatiotemporal attention to relate local regions without explicit point tracking \cite{9740525}.

Recent work extends this progression toward longer sequences, structured geometry, and efficient sequence modeling. Point Primitive Transformer represents long-term point cloud videos through primitive planes and hierarchical attention \cite{10.1007/978-3-031-19818-2_2}, while LeaF learns local coordinate frames to factorize geometry and motion under changing poses \cite{10377208}. State-space backbones such as Mamba4D and UST-SSM reorganize 4D point data into sequences for efficient long-range modeling \cite{Liu_2025_CVPR,Li_2025_ICCV}. Spectral approaches further mix spatial, temporal, and frequency-domain cues for point cloud action recognition and 4D understanding \cite{li2026stsmixerspatiotemporalspectralmixer4d,11533509}. These supervised methods provide increasingly capable 4D backbones, but their representations are still primarily shaped by labeled downstream objectives.

\subsection{Self-Supervised Learning on Point Cloud Videos}
\label{subsec:related-ssl-point-cloud-video}

Self-supervised learning on point cloud videos reduces reliance on costly sequence-level annotations by deriving training targets from the 4D signal itself. Early work used temporal pretext tasks, such as predicting the order of shuffled point cloud clips, to encourage motion-sensitive representations \cite{9423387}. Contrastive objectives construct positive and negative relations in feature space: PointCMP combines local and global branches with contrastive mask prediction \cite{Shen_2023_CVPR}, point contrastive prediction with semantic clustering performs point-level contrast over superpoints \cite{10376612}, and contrastive predictive autoencoders combine prediction, contrastive learning, and reconstruction for dynamic point cloud sequences \cite{Sheng_Shen_Xiao_2023}. Distillation-based methods provide another route, with Complete-to-Partial 4D Distillation training a student from partial observations under guidance from complete sequences \cite{Zhang_2023_CVPR}.

Masked modeling is another central direction. Image and video MAE methods show that high-ratio masking with reconstruction can produce transferable visual representations \cite{He_2022_CVPR,NEURIPS2022_e97d1081}, and static point cloud pretraining adapts this idea through point tokens or masked point patches \cite{9880161,10.1007/978-3-031-20086-1_35}. For point cloud videos, MaST-Pre masks spatiotemporal point tubes and couples point-tube reconstruction with temporal cardinality-difference prediction \cite{10378457}. Later methods blend masking with additional supervision: M2PSC adds masked motion trajectory prediction and semantic contrast \cite{10.1007/978-3-031-73116-7_24}, while Uni4D disentangles low-level geometry reconstruction from latent alignment of frame-level motion and video-level information \cite{zuo2025uni4dunifiedselfsupervisedlearning}. More recently, DiMP introduces diffusion-based masked-center inference and probabilistic inter-frame motion supervision while retaining Chamfer-based geometric reconstruction \cite{zhang2026diffusionmaskedpretrainingdynamic}. These methods establish the usefulness of unlabeled 4D data, but many still depend on coordinate reconstruction, explicit motion proxy targets, contrastive sample design, or teacher targets.

\subsection{Joint-Embedding Predictive Architectures}
\label{subsec:related-jepa}

Joint-Embedding Predictive Architectures learn by predicting target representations from context representations rather than reconstructing raw observations. This non-generative formulation encourages the encoder to model information useful at the latent semantic level, reducing pressure to preserve all low-level details. I-JEPA applies this idea to images by predicting target-block embeddings from visible context blocks \cite{10205476}, and V-JEPA extends feature prediction to video using latent targets without reconstruction, negative examples, text supervision, or pretrained image encoders \cite{bardes2024revisiting}. Point-JEPA and 3D-JEPA adapt the same principle to point clouds through context-target sampling over point patches or 3D blocks \cite{10943937,hu20243djepajointembeddingpredictive}. At the 4D point-cloud level, Cross4D-JEPA extends latent-space matching to 4D point clouds by distilling dense per-point targets from frozen 2D or video foundation models into a 4D point encoder \cite{nguyen2026cross4djepadensecrossmodalcorrespondence}.

A central challenge in JEPA-style learning is representation collapse. Earlier systems commonly stabilize training with a target encoder or teacher branch, often combined with stop-gradient and exponential moving average updates \cite{10205476,bardes2024revisiting}. Recent work moves toward explicit embedding regularization. LeJEPA introduces Sketched Isotropic Gaussian Regularization (SIGReg), which constrains embeddings toward an isotropic Gaussian distribution and reduces reliance on teacher-student encoders, stop-gradient operations, and related heuristics \cite{balestriero2025lejepaprovablescalableselfsupervised}. LeWorldModel applies this direction to end-to-end latent world modeling from pixels using a predictive loss together with Gaussian latent regularization \cite{maes2026leworldmodelstableendtoendjointembedding}. This transition motivates a simpler JEPA-style formulation for 4D point cloud videos, where contextual features are used to predict target representations in latent space rather than reconstructing the input.

\section{Method}
\label{sec:method}

\begin{figure}[!t]
    \centering
    \includegraphics[width=\linewidth]{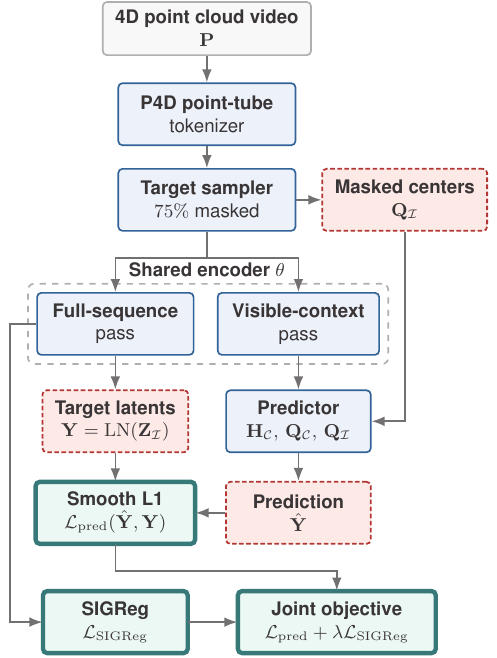}
    \caption{Overview of the proposed pretraining framework. Point-tube tokens are split into visible context and masked targets; a predictor infers target latents while SIGReg regularizes the latent distribution.}
    \label{fig:overall-framework}
\end{figure}

The proposed pretraining framework learns from unlabeled 4D point cloud videos by predicting masked spatiotemporal token representations in latent space. Given an input clip $\mathbf{P}\in\mathbb{R}^{T\times N\times 3}$, the model first converts local space-time neighborhoods into point-tube tokens, hides a large subset of these tokens from the context encoder, and trains a predictor to infer their latent representations. In contrast to masked autoencoding methods that reconstruct point coordinates or auxiliary motion targets, the objective is defined on encoder features and is regularized by SIGReg to discourage collapse. Figure~\ref{fig:overall-framework} summarizes the encoding pipeline, predictor, and training objective.

\subsection{Tokenization and Masking}
\label{subsec:method-tokenization-masking}

Point cloud videos are irregular in both space and time, so masking individual raw points would not provide stable semantic units. We therefore adopt the point-tube tokenization used in P4Transformer and prior masked point-cloud-video pretraining~\cite{9578674,10378457,zuo2025uni4dunifiedselfsupervisedlearning}. Point-tube centers are sampled from $\mathbf{P}$ using Farthest Point Sampling (FPS); for each center $\hat{p}_i$, we define $\boldsymbol{Tube}_{\hat{p}_i}=\{p\mid p\in\mathbf{P}, D_s(p,\hat{p}_i)<r_s, D_t(p,\hat{p}_i)\leq \frac{r_t}{2}\}$. The tokenizer converts local spatiotemporal neighborhoods into a sequence of point-tube embeddings $\mathbf{X}=\{e_i\}_{i=1}^{S}$, which serve as the units for masking and latent prediction.

Each token is associated with a 4D center $\mathbf{Q}=\{q_i\}_{i=1}^{S}$ indicating its spatial anchor and temporal position. These centers are used as positional information for the encoder and predictor, allowing the model to distinguish visible and masked point tubes by their locations. Since the prediction target is defined in latent space, the framework does not require the tokenizer to support raw-coordinate reconstruction.

Masking is performed after tokenization at the point-tube-token level. A target sampler selects a target index set $\mathcal{I}$ from the token sequence. We use random target sampling with a high mask ratio of $75\%$, consistent with the observation in masked image and video modeling that high masking ratios reduce shortcut reconstruction and create a stronger prediction task~\cite{He_2022_CVPR,NEURIPS2022_e97d1081,10378457}. The context set is then formed by excluding the target indices,
\begin{equation}
    \mathcal{C}=\{1,\ldots,S\}\setminus\mathcal{I}.
\end{equation}
In the pretraining configuration, the context encoder receives the remaining non-target tokens. Consequently, the model is trained to infer masked point-tube representations from the visible complement of the same 4D point cloud video.

\subsection{Architecture}
\label{subsec:method-architecture}

The architecture consists of a P4D tokenizer, a Transformer encoder, and a lightweight predictor. This design follows the asymmetric spirit of masked modeling, where prediction is conditioned only on visible tokens, but replaces coordinate reconstruction with JEPA-style latent prediction~\cite{10205476,bardes2024revisiting,10943937}. Compared with MAE-style point-cloud-video pretraining such as MaST-Pre and Uni4D, the predictor is not a geometric decoder and does not output point coordinates, Chamfer targets, or hand-crafted motion descriptors~\cite{10378457,zuo2025uni4dunifiedselfsupervisedlearning}.

\paragraph{Encoder.}
The encoder maps point-tube tokens and their 4D positional information to latent spatiotemporal representations. During pretraining, the same encoder is used in two passes. A full-sequence pass encodes all tokens and produces latents $\mathbf{Z}$. A context pass encodes only the visible tokens $\mathbf{X}_{\mathcal{C}}$ with their corresponding centers $\mathbf{Q}_{\mathcal{C}}$, producing context features $\mathbf{H}_{\mathcal{C}}$ for the predictor. The SIGReg pretraining path therefore does not use a separate EMA teacher; the target representation and context representation are produced by the shared encoder.

\paragraph{Predictor.}
The predictor receives the context features, the context centers, and the masked target centers. Context features are first projected to a lower predictor dimension. For the target index set $\mathcal{I}$, the predictor appends learned mask-token slots whose positional embeddings are computed from $\mathbf{Q}_{\mathcal{I}}$. The concatenated context tokens and target query slots are processed by the predictor network. Only the outputs at the target slots are kept and projected back to the encoder dimension:
\begin{equation}
    \hat{\mathbf{Y}}
    =
    h_{\phi}
    \left(
        \mathbf{H}_{\mathcal{C}},
        \mathbf{Q}_{\mathcal{C}},
        \mathbf{Q}_{\mathcal{I}}
    \right).
\end{equation}
This formulation gives the predictor both the visible latent context and the 4D locations of the missing point tubes, while preventing it from observing the masked token features directly.

\subsection{Training Objective}
\label{subsec:method-training-objective}

The training objective combines masked latent prediction with an anti-collapse regularizer:
\begin{equation}
    \mathcal{L}
    =
    \mathcal{L}_{\mathrm{pred}}
    +
    \lambda\,\mathcal{L}_{\mathrm{SIGReg}}.
    \label{eq:overall-loss}
\end{equation}

\paragraph{Prediction loss.}
For the sampled target set $\mathcal{I}$, the full-sequence encoder first produces latent representations $\mathbf{Z}$, and the prediction targets are formed by applying feature-wise layer normalization to the selected masked latents, $\mathbf{Y}=\mathrm{LN}(\mathbf{Z}_{\mathcal{I}})$. The predictor outputs $\hat{\mathbf{Y}}$ from the visible context. The latent prediction loss is
\begin{equation}
    \mathcal{L}_{\mathrm{pred}}
    =
    \mathcal{L}_{\mathrm{smooth\text{-}L1}}
    \left(
        \hat{\mathbf{Y}},
        \mathbf{Y}
    \right).
    \label{eq:prediction-loss}
\end{equation}
Since both arguments of Eq.~\ref{eq:prediction-loss} are latent token representations, this objective encourages the model to capture predictable spatiotemporal semantics rather than recover the raw input points exactly.

\begin{figure}[!t]
    \centering
    \includegraphics[width=\linewidth]{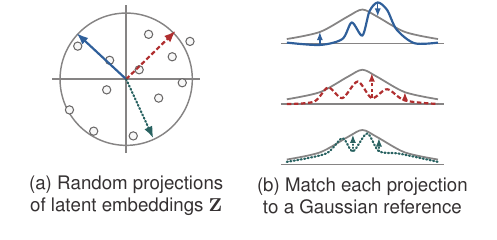}
    \caption{Illustration of SIGReg. Random one-dimensional projections are drawn from the latent embedding distribution, and each projected distribution is matched to a Gaussian reference.}
    \label{fig:sigreg-regularization}
\end{figure}

\paragraph{SIGReg.}
Latent prediction alone can admit collapsed solutions in which the encoder produces low-variance features that are easy to predict. To stabilize training without a momentum teacher or stop-gradient target branch, we apply Sketched Isotropic Gaussian Regularization (SIGReg) to the full-sequence latents~\cite{balestriero2025lejepaprovablescalableselfsupervised,maes2026leworldmodelstableendtoendjointembedding}. Let \(\mathbf{Z}\in\mathbb{R}^{S\times B\times d}\) denote the full-sequence encoder representations. SIGReg draws random unit projection directions $u_j$ and matches the one-dimensional projected distributions $\mathbf{Z}u_j$ to an isotropic Gaussian reference using a sketched characteristic-function statistic:
\begin{equation}
    \mathcal{L}_{\mathrm{SIGReg}}
    =
    \frac{1}{J}
    \sum_{j=1}^{J}
    \mathcal{T}(\mathbf{Z}u_j),
    \label{eq:sigreg-loss}
\end{equation}
where $\mathcal{T}(\cdot)$ denotes the Epps--Pulley-style statistic used by SIGReg. Its explicit finite-sample form and implementation details are provided in the supplementary material. The final loss in Eq.~\ref{eq:overall-loss} therefore balances two complementary pressures: the predictor must infer missing point-tube latents from visible 4D context, while the encoder is regularized to maintain a non-degenerate and approximately isotropic latent space. Figure~\ref{fig:sigreg-regularization} illustrates this projection-and-matching view of the regularizer.

\section{Experiments}
\label{sec:experiments}

We conduct experiments on MSRAction-3D~\cite{li2010actionrecognitionbag3dpoints} and NTU RGB+D~\cite{shahroudy2016ntu} for action recognition, and on SHREC'17~\cite{10.2312:3dor.20171049} for cross-dataset gesture-recognition transfer. Implementation details, including the computing environment and hyperparameter settings, are provided in the supplementary material.

\subsection{Pretraining}
\label{subsec:experiments-pretraining}

We pretrain the proposed JEPA-style model on the training split defined by the cross-subject protocol of MSRAction-3D~\cite{li2010actionrecognitionbag3dpoints} and NTU RGB+D~\cite{shahroudy2016ntu}. Following P4Transformer-based protocols used in prior point-cloud-video studies~\cite{9578674,10378457,10.1007/978-3-031-73116-7_24,zuo2025uni4dunifiedselfsupervisedlearning}, each pretraining clip contains 24 frames with 1024 sampled points per frame. The frame sampling stride is set to 1 for MSRAction-3D and 2 for NTU RGB+D, and random scaling is applied as data augmentation.

P4Transformer is used as the encoder, with 5 layers on MSRAction-3D and 10 layers on NTU RGB+D, matching common practice for these benchmarks~\cite{9578674,10378457,zuo2025uni4dunifiedselfsupervisedlearning}. The pretraining task predicts masked point-tube representations from the visible context, using Smooth L1 prediction loss together with SIGReg. We optimize with AdamW and cosine learning-rate decay. The model is pretrained for 200 epochs on MSRAction-3D and for 100 epochs on NTU RGB+D.

\subsection{End-to-End Fine-Tuning}
\label{subsec:experiments-finetuning}

\subsubsection{MSRAction-3D}
\label{subsec:experiments-finetuning-msr}

For MSRAction-3D, we initialize the P4Transformer encoder from the self-supervised checkpoint and replace the pretraining predictor with a supervised classification head. Following prior fine-tuning settings~\cite{9578674,10378457}, each clip contains 24 frames and 2048 points per frame. The model is trained end-to-end for 50 epochs with AdamW, cosine learning-rate decay, and cross-entropy loss.

Table~\ref{tab:msr-action3d} shows that self-supervised pretraining improves the P4Transformer backbone and gives the strongest result among the listed P4Transformer-based methods. This suggests that latent prediction with SIGReg provides useful spatiotemporal initialization even on the relatively small MSRAction-3D benchmark.

\begin{table}[!t]
    \centering
    \small
    \setlength{\tabcolsep}{2pt}
    \begin{tabular}{@{}>{\raggedright\arraybackslash}p{0.73\linewidth}c@{}}
        \toprule
        Method & Acc. \\
        \midrule
        \multicolumn{2}{c}{\textit{Supervised Learning Only}} \\
        \midrule
        MeteorNet~\cite{9010250} & 88.50 \\
        PSTNet~\cite{fan2021pstnet} & 91.20 \\
        PSTNet++~\cite{9650574} & 92.68 \\
        Kinet~\cite{Zhong_2022_CVPR} & 93.27 \\
        PPTr~\cite{10.1007/978-3-031-19818-2_2} & 92.33 \\
        3DInAction~\cite{Ben-Shabat_2024_CVPR} & 92.23 \\
        Mamba4D~\cite{Liu_2025_CVPR} & 92.68 \\
        P4Transformer~\cite{9578674} & 90.94 \\
        \midrule
        \multicolumn{2}{c}{\textit{With Self-Supervised Pretraining}} \\
        \midrule
        P4Transformer + MaST-Pre~\cite{10378457} & 91.29 \\
        P4Transformer + M2PSC~\cite{10.1007/978-3-031-73116-7_24} & 93.03 \\
        P4Transformer + Uni4D~\cite{zuo2025uni4dunifiedselfsupervisedlearning} & 93.38 \\
        P4Transformer + DiMP~\cite{zhang2026diffusionmaskedpretrainingdynamic} & 93.97 \\
        \midrule
        P4Transformer + \textbf{Ours} & \textbf{94.08} \\
        \bottomrule
    \end{tabular}
    \caption{Action recognition accuracy (\%) on MSRAction-3D. Baselines are grouped by training setting.}
    \label{tab:msr-action3d}
\end{table}

\subsubsection{NTU RGB+D}
\label{subsec:experiments-finetuning-ntu}

For NTU RGB+D, we follow the cross-subject protocol. Fine-tuning uses 24-frame clips and 2048 sampled points per frame. The model is fine-tuned for 20 epochs with AdamW, cosine learning-rate decay, and cross-entropy loss.

The full-label column of Table~\ref{tab:ntu-rgbd-semi} reports the end-to-end fine-tuning result. Our model improves over the supervised P4Transformer baseline and also outperforms the listed self-supervised baselines. Moreover, this result is obtained with a 100-epoch pretraining schedule on NTU RGB+D, while MaST-Pre and M2PSC report 200 epochs~\cite{10378457,10.1007/978-3-031-73116-7_24}.

\begin{table}[!t]
    \centering
    \small
    \setlength{\tabcolsep}{2pt}
    \begin{tabular}{@{}>{\raggedright\arraybackslash}p{0.50\linewidth}cc@{}}
        \toprule
        Method & Full & 50\% Labels \\
        \midrule
        \multicolumn{3}{c}{\textit{Supervised Learning Only}} \\
        \midrule
        3DV-Motion~\cite{a7f3d18e98504c43854f266517083e26} & 84.5 & -- \\
        3DV-PointNet++~\cite{a7f3d18e98504c43854f266517083e26} & 88.8 & -- \\
        PSTNet~\cite{fan2021pstnet} & 90.5 & -- \\
        PSTNet++~\cite{9650574} & 91.4 & -- \\
        Kinet~\cite{Zhong_2022_CVPR} & 92.3 & -- \\
        P4Transformer~\cite{9578674} & 90.2 & 81.2 \\
        \midrule
        \multicolumn{3}{c}{\textit{With Self-Supervised Pretraining}} \\
        \midrule
        P4Transformer + MaST-Pre~\cite{10378457} & 90.8 & 87.8 \\
        P4Transformer + M2PSC~\cite{10.1007/978-3-031-73116-7_24} & 91.3 & 88.7 \\
        P4Transformer + Uni4D~\cite{zuo2025uni4dunifiedselfsupervisedlearning} & 90.7 & 86.5 \\
        \midrule
        P4Transformer + \textbf{Ours} & \textbf{91.8} & \textbf{89.0} \\
        \bottomrule
    \end{tabular}
    \caption{Action recognition accuracy (\%) on NTU RGB+D under the cross-subject protocol. The semi-supervised setting fine-tunes with 50\% labeled training videos.}
    \label{tab:ntu-rgbd-semi}
\end{table}

\subsection{Semi-Supervised Learning}
\label{subsec:experiments-semi}

We further evaluate data efficiency on NTU RGB+D. Following the semi-supervised evaluation protocol used in prior point-cloud-video representation learning studies~\cite{10378457,10.1007/978-3-031-73116-7_24,zuo2025uni4dunifiedselfsupervisedlearning}, the encoder is first pretrained on the full unlabeled cross-subject training split, and supervised fine-tuning then uses only 50\% of the labeled training videos. The test split and evaluation protocol remain unchanged, and all other settings follow the NTU end-to-end fine-tuning setup.

As shown under the 50\% Labels setting in Table~\ref{tab:ntu-rgbd-semi}, pretraining gives a substantially larger improvement under limited labels than in the full-label setting. This indicates that the learned latent representation is especially useful when downstream supervision is scarce.

\subsection{Few-Shot Learning}
\label{subsec:experiments-few-shot}

Few-shot learning is evaluated on MSRAction-3D using the all-class protocol in Table~\ref{tab:few-shot-msr}. We initialize the encoder with weights from self-supervised pretraining on NTU RGB+D, append a lightweight classifier, and fine-tune the full model on MSRAction-3D. This setting differs from the $n$-way $m$-shot protocol used by Uni4D~\cite{zuo2025uni4dunifiedselfsupervisedlearning}, where only a subset of action classes is sampled for each few-shot setting. In our evaluation, all action classes are retained, but ``shot'' still denotes the number of labeled training videos sampled per class, set to 1, 3, or 5. To reduce sensitivity to support-set sampling, we report the mean and standard deviation over ten runs. For each run, the scratch and pretrained models use the same support set, allowing a paired comparison under identical labeled examples.

\begin{table}[!t]
    \centering
    \small
    \setlength{\tabcolsep}{3pt}
    \begin{tabular}{@{}>{\raggedright\arraybackslash}p{0.26\linewidth}ccc@{}}
        \toprule
        Initialization & 1-shot & 3-shot & 5-shot \\
        \midrule
        Scratch & $20.07{\pm}10.40$ & $49.79{\pm}22.35$ & $75.89{\pm}10.32$ \\
        Pretrained & $\mathbf{31.05{\pm}9.15}$ & $\mathbf{63.41{\pm}5.43}$ & $\mathbf{79.76{\pm}4.07}$ \\
        \midrule
        Gain & $+10.98$ & $+13.62$ & $+3.87$ \\
        \bottomrule
    \end{tabular}
    \caption{All-class few-shot P4Transformer video-level top-1 action recognition accuracy (\%) on MSRAction-3D. Results are reported as mean $\pm$ standard deviation over ten paired support-set runs.}
    \label{tab:few-shot-msr}
\end{table}

The proposed pretraining improves the mean accuracy over training from scratch for all three shot budgets, with the largest gains in the 1-shot and 3-shot settings. Pretraining also substantially reduces the standard deviation in the 3-shot and 5-shot settings, indicating that the NTU RGB+D representation makes MSRAction-3D adaptation less dependent on a favorable support-set draw. The smaller 5-shot gain is expected because additional labeled examples reduce the relative value of initialization.

\subsection{Transfer Learning}
\label{subsec:experiments-transfer}

We evaluate transfer learning by following prior cross-dataset protocols~\cite{10378457,zuo2025uni4dunifiedselfsupervisedlearning}. The encoder is pretrained on NTU RGB+D and then fine-tuned on SHREC'17~\cite{10.2312:3dor.20171049} under the 28-class setting, using a newly initialized classifier. This setting transfers from action recognition to hand-gesture recognition, testing both cross-dataset and cross-task generalization.

As shown in Table~\ref{tab:transfer-learning}, NTU RGB+D pretraining improves P4Transformer on SHREC'17 at both the 30- and 50-epoch evaluation points. The relative improvement over the corresponding P4Transformer baseline is larger with fewer fine-tuning epochs, suggesting that the learned representation provides a useful initialization for faster adaptation.

\begin{table}[!t]
    \centering
    \small
    \setlength{\tabcolsep}{2pt}
    \begin{tabular}{@{}>{\raggedright\arraybackslash}p{0.52\linewidth}cc@{}}
        \toprule
        Method & Epochs & Acc. \\
        \midrule
        \multicolumn{3}{c}{\textit{Supervised Learning Only}} \\
        \midrule
        PointLSTM~\cite{9157795} & -- & 87.6 \\
        PointLSTM-PSS~\cite{9157795} & -- & 93.1 \\
        Kinet~\cite{Zhong_2022_CVPR} & -- & 95.2 \\
        P4Transformer~\cite{9578674} & 30 & 87.5 \\
        P4Transformer~\cite{9578674} & 50 & 91.2 \\
        \midrule
        \multicolumn{3}{c}{\textit{With Self-Supervised Pretraining}} \\
        \midrule
        P4Transformer + MaST-Pre~\cite{10378457} & 30 & 90.2 \\
        P4Transformer + MaST-Pre~\cite{10378457} & 50 & 92.4 \\
        P4Transformer + M2PSC~\cite{10.1007/978-3-031-73116-7_24} & 30 & 90.9 \\
        P4Transformer + M2PSC~\cite{10.1007/978-3-031-73116-7_24} & 50 & 92.8 \\
        P4Transformer + Uni4D~\cite{zuo2025uni4dunifiedselfsupervisedlearning} & 50 & 93.8 \\
        P4Transformer + \textbf{Ours} & 30 & \textbf{91.7} \\
        P4Transformer + \textbf{Ours} & 50 & \textbf{93.3} \\
        \bottomrule
    \end{tabular}
    \caption{Transfer learning accuracy (\%) on SHREC'17.}
    \label{tab:transfer-learning}
\end{table}

\subsection{Visualization}
\label{subsec:experiments-visualization}

\begin{figure*}[!t]
    \centering
    \includegraphics[width=\textwidth]{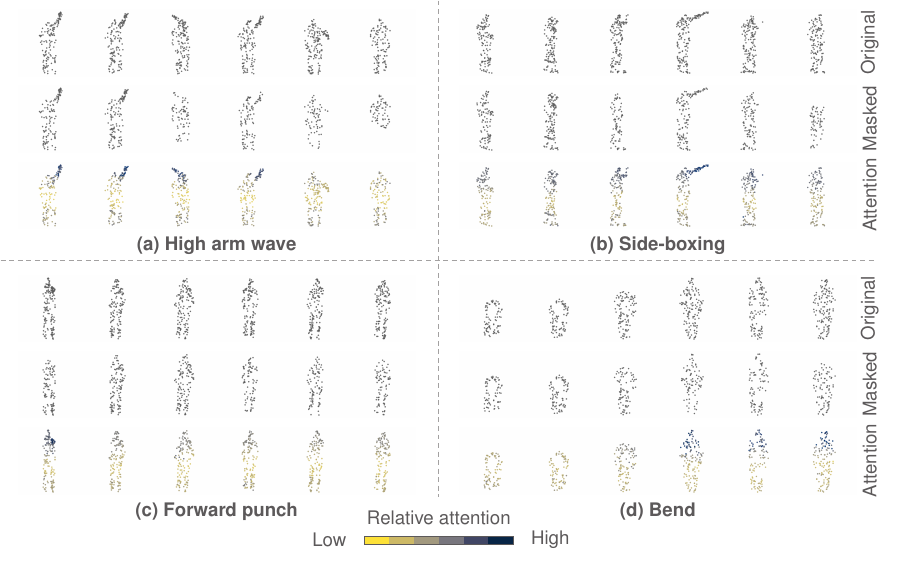}
    \caption{Qualitative visualization of attention maps from the MSR-pretrained encoder. Each panel contains the original point cloud sequence, the masked point cloud sequence, and the encoder attention map from top to bottom.}
    \label{fig:attention-visualization}
\end{figure*}

To qualitatively examine the representation learned by self-supervised pretraining, we visualize point-cloud clips from MSRAction-3D together with their masked inputs and attention maps obtained from the MSR-pretrained encoder. The attention map is computed directly from the encoder self-attention weights. Specifically, we use the last encoder layer, average the attention assigned to each point-tube token over all heads and all query tokens, and obtain one scalar score for each tube token. This token-level score is then projected back to the point domain by assigning it to the neighboring points contained in the corresponding tube. In each panel of Figure~\ref{fig:attention-visualization}, the first row shows the original point cloud sequence, the second row shows the masked point cloud, and the third row shows the corresponding encoder attention map.

The attention maps are concentrated on semantically relevant body parts for different actions. For high arm wave and side-boxing, the strongest responses appear around the arms; for forward punch, they move toward the punching hand; and for bend, they are concentrated near the head and upper body. Although this visualization is qualitative, the consistency between the attended regions and the action-defining motion cues suggests that MSR pretraining helps the encoder emphasize informative spatiotemporal regions rather than only low-level geometric structure.

\subsection{Ablation Studies}
\label{subsec:experiments-ablation}

We perform controlled ablations on MSRAction-3D to isolate the target path, target-position query, masking ratio, and SIGReg strength. All variants follow the MSRAction-3D pretraining and fine-tuning protocols described above.

\begin{figure}[!t]
    \centering
    \includegraphics[width=0.9\linewidth]{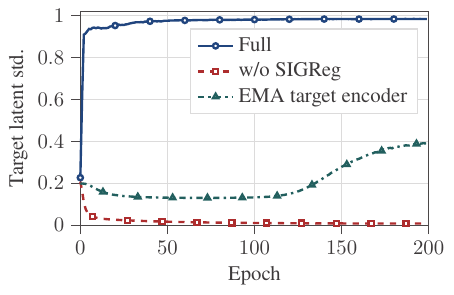}
    \caption{Target-latent standard deviation over pretraining epochs for representative component-ablation variants on MSRAction-3D.}
    \label{fig:target-latent-std}
\end{figure}

\paragraph{Component ablations.}
Table~\ref{tab:component-ablation} studies these training-path and predictor-design choices. Removing SIGReg lowers accuracy to 89.90\%, and Figure~\ref{fig:target-latent-std} shows that this degradation is accompanied by early collapse of the target-latent standard deviation. The full model rapidly stabilizes near unit target-latent standard deviation, whereas the EMA target encoder remains substantially lower for most of pretraining and also underperforms the shared-encoder SIGReg formulation. The stop-gradient target variant, defined by detaching the layer-normalized target latents $\mathbf{Y}$ used in Eq.~\ref{eq:prediction-loss}, has the largest downstream drop, suggesting that detaching target latents is harmful in this non-EMA training path. Removing the target-position query causes a modest accuracy drop from 94.08\% to 93.03\%. Here, the ablated target slots use only the shared learned mask token, without the positional embedding computed from the masked centers $\mathbf{Q}_{\mathcal{I}}$. This result indicates that visible context alone provides substantial predictive information on this benchmark, while explicit target-position queries still provide a measurable benefit.

\begin{table}[!t]
    \centering
    \small
    \setlength{\tabcolsep}{4pt}
    \begin{tabular}{@{}>{\raggedright\arraybackslash}p{0.70\linewidth}c@{}}
        \toprule
        Variant & Acc. \\
        \midrule
        Full model & \textbf{94.08} \\
        w/o SIGReg & 89.90 \\
        Stop-gradient targets & 79.44 \\
        w/o target-position query & 93.03 \\
        EMA target encoder & 88.50 \\
        \bottomrule
    \end{tabular}
    \caption{Component/path ablation on MSRAction-3D.}
    \label{tab:component-ablation}
\end{table}

\paragraph{Sensitivity ablations.}
Table~\ref{tab:sensitivity-ablation} summarizes the remaining sensitivity to target masking and SIGReg strength. For masking, the 75\% ratio gives the best accuracy among the tested values. A lower ratio provides a weaker prediction task because more context tokens remain visible, whereas an 85\% ratio removes too much context for reliable target inference on MSRAction-3D. Since mask-ratio variants maintain near-unit target-latent standard deviation, Table~\ref{tab:sensitivity-ablation} reports their accuracy only; for SIGReg-strength variants, Target Latent Std. denotes the target-latent standard deviation averaged over the final ten pretraining epochs. Among SIGReg weights, $\lambda_{\mathrm{SIGReg}}=0.01$ performs best; a smaller weight yields lower target-latent standard deviation and weaker downstream accuracy, while a larger weight maintains high variance but can overemphasize distribution matching relative to predictive alignment.

\begin{table}[!t]
    \centering
    \small
    \setlength{\tabcolsep}{5pt}
    \begin{tabular}{@{}>{\raggedright\arraybackslash}p{0.36\linewidth}ccc@{}}
        \toprule
        Setting & Value & Target Latent Std. & Acc. \\
        \midrule
        Mask ratio & 0.65 & -- & 88.85 \\
        Mask ratio & 0.75 & -- & \textbf{94.08} \\
        Mask ratio & 0.85 & -- & 91.99 \\
        \midrule
        $\lambda_{\mathrm{SIGReg}}$ & 0.001 & 0.870 & 86.76 \\
        $\lambda_{\mathrm{SIGReg}}$ & 0.01 & 0.984 & \textbf{94.08} \\
        $\lambda_{\mathrm{SIGReg}}$ & 0.1 & 0.992 & 89.55 \\
        \bottomrule
    \end{tabular}
    \caption{Sensitivity ablations on MSRAction-3D.}
    \label{tab:sensitivity-ablation}
\end{table}

\section{Conclusion}
\label{sec:conclusion}

In this work, we studied JEPA-style self-supervised learning for 4D point cloud videos. The proposed framework converts point cloud sequences into spatiotemporal point-tube tokens, masks a large subset of these tokens, and trains a predictor to infer masked target representations from visible context representations in latent space. Instead of reconstructing raw point coordinates, the model learns through feature-space prediction, while SIGReg regularizes the learned latent distribution to reduce representation collapse in the proposed shared-encoder training path.

We instantiate this formulation with a P4D tokenizer, Transformer encoder, and lightweight predictor, and evaluate the pretrained encoder across multiple downstream, transfer, qualitative, and ablation settings. The results show that latent prediction with SIGReg provides a strong initialization for 4D action and gesture recognition, especially under limited labeled data. This study remains limited to P4Transformer-based backbones and a modest benchmark scope, motivating future work on larger-scale pretraining, broader 4D datasets, and alternative architectures.

\bibliography{aaai2027}

\appendix
\twocolumn[
    \begin{center}
        {\LARGE\bfseries Joint-Embedding Prediction of Masked Point Tubes for Self-Supervised Learning on 4D Point Cloud Videos\par}
        \vspace{0.5em}
        {\Large\bfseries Supplementary Material\par}
    \end{center}
    \vspace{0.75em}
]

\section{Overview}
\label{sec:supp_overview}

This supplementary document records implementation details and additional
diagnostics that support the main paper. It first gives the
finite-sample SIGReg objective used in pretraining, including the latent tensor
shape, random projections, characteristic-function statistic, and gradient path.
It then reports the experimental environment, data protocols, training settings,
random seeds, and evaluation procedure. The final part provides
ablation diagnostics and the per-seed few-shot results summarized in the main
paper.

\section{SIGReg Details}
\label{sec:supp_sigreg}

\subsection{Regularized Representation}

Let $\mathbf{Z}\in\mathbb{R}^{S\times B\times d}$ denote the full-sequence
encoder latents used by SIGReg, where $S$ is the number of point-tube tokens,
$B$ is the minibatch size, and $d$ is the encoder dimension. In the
implementation, these latents are the last hidden states of the full-sequence
encoder pass and are supplied to the regularizer with shape $(S,B,d)$.

\subsection{Random Projections}

At each forward pass, SIGReg samples a projection matrix
$\mathbf{A}=[u_1,\ldots,u_J]\in\mathbb{R}^{d\times J}$ from a standard normal
distribution and normalizes each column to unit Euclidean norm. For token
position $i$, batch element $b$, and projection $j$, define the scalar
projection
\begin{equation}
    r_{i,b,j} = \mathbf{z}_{i,b}^{\top}u_j .
\end{equation}
The implementation uses $J=1024$ random projections. These directions
are resampled at every forward pass and are not trainable parameters.

\subsection{Characteristic-Function Matching}

The one-dimensional projected empirical distribution is matched to the
standard Gaussian characteristic function through a truncated Epps--Pulley
discrepancy. For token position $i$ and projection $j$, let
$\hat{\psi}_{i,j}(t)=B^{-1}\sum_{b=1}^{B}
\exp\left(\mathrm{i}t r_{i,b,j}\right)$ denote the empirical characteristic
function. The continuous-form quantity is
\begin{equation}
    \mathcal{D}_{i,j}
    =
    B\int_{-3}^{3} e^{-t^2/2}
    \left\lvert
        \hat{\psi}_{i,j}(t)-e^{-t^2/2}
    \right\rvert^2\,dt .
\end{equation}
The implementation approximates this integral on a fixed frequency grid. With
$K=17$ knots, define
\begin{equation}
    \tau_k = \frac{3(k-1)}{K-1},\qquad
    \phi_k = \exp\left(-\frac{\tau_k^2}{2}\right),
\end{equation}
for $k=1,\ldots,K$. Let $\Delta=3/(K-1)$. Since the integrand is even
in the frequency variable, we evaluate only the nonnegative half of
$[-3,3]$ and fold the negative-frequency contribution into the quadrature
weights. The implementation uses the following quadrature coefficients:
\begin{equation}
    \omega_k =
    \begin{cases}
        \Delta, & k=1 \text{ or } k=K,\\
        2\Delta, & 1<k<K .
    \end{cases}
\end{equation}
For each token position and projection, the empirical real and imaginary
parts of the characteristic function are
\begin{equation}
    \begin{aligned}
    \hat{c}_{i,j,k}
    =
    \frac{1}{B}\sum_{b=1}^{B}\cos(\tau_k r_{i,b,j}),
    \\
    \hat{s}_{i,j,k}
    =
    \frac{1}{B}\sum_{b=1}^{B}\sin(\tau_k r_{i,b,j}).
    \end{aligned}
\end{equation}
Since a standard Gaussian has characteristic function $\phi_k$ and zero
imaginary component, the per-token, per-projection statistic is
\begin{equation}
    \mathcal{T}_{i,j}
    =
    B
    \sum_{k=1}^{K}
    \omega_k\phi_k
    \left[
        \left(\hat{c}_{i,j,k}-\phi_k\right)^2
        +
        \hat{s}_{i,j,k}^{2}
    \right].
    \label{eq:supp-sigreg-statistic}
\end{equation}
The final SIGReg term first averages token-level statistics within each random
projection and then averages over projections:
\begin{equation}
    \mathcal{L}_{\mathrm{SIGReg}}
    =
    \frac{1}{J}
    \sum_{j=1}^{J}
    \left(
    \frac{1}{S}
    \sum_{i=1}^{S}
    \mathcal{T}_{i,j}
    \right).
    \label{eq:supp-sigreg-loss}
\end{equation}
The multiplicative factor $B$ follows the finite-sample
Epps--Pulley statistic used by SIGReg and is retained in the implementation.
Additionally, the fixed frequency grid $\tau_k$, Gaussian reference values
$\phi_k$, and combined weights $\omega_k\phi_k$ are precomputed and stored
as non-trainable buffers.

\subsection{Training Objective}

During pretraining, the model first performs a full-sequence encoder pass.
Masked prediction targets are formed by applying feature-wise layer normalization
to the selected target latents. The SIGReg loss is computed separately on the full set of last hidden states from the same full-sequence pass. The
optimization objective is
\begin{equation}
    \mathcal{L}
    =
    \mathcal{L}_{\mathrm{pred}}
    +
    \lambda_{\mathrm{SIGReg}}\mathcal{L}_{\mathrm{SIGReg}},
\end{equation}
where $\lambda_{\mathrm{SIGReg}}$ is the configured regularization weight. The
default training path does not detach the prediction targets: gradients from
both $\mathcal{L}_{\mathrm{pred}}$ and $\mathcal{L}_{\mathrm{SIGReg}}$ can update
the shared encoder. The stop-gradient and EMA variants reported in the main
paper are ablations of this default path.

\section{Implementation Details}
\label{sec:supp_implementation}

\subsection{Infrastructure}

All experiments were executed on a workstation using one CUDA device per
training run. The computing infrastructure and software environment were as
follows:
\begin{itemize}
    \item \textbf{OS}: Rocky Linux 9.8.
    \item \textbf{CPU}: 2$\times$ Intel Xeon Gold 5420+.
    \item \textbf{System memory}: 384 GiB.
    \item \textbf{GPU}: 4$\times$ NVIDIA RTX 6000 Ada Generation.
    \item \textbf{GPU driver}: NVIDIA 580.159.04.
    \item \textbf{Python and GPU stack}: Python 3.10.19; PyTorch 2.6.0+cu124; PyTorch CUDA build 12.4; cuDNN 9.1.0.
    \item \textbf{Scientific libraries}: NumPy 2.2.5; SciPy 1.15.3; scikit-learn 1.7.2.
    \item \textbf{Utility libraries}: torchvision 0.21.0+cu124; tqdm 4.65.2.
\end{itemize}

\subsection{Data Protocols}

Table~\ref{tab:supp-data-protocols} summarizes the dataset-level class counts,
video-level split sizes, and cross-subject split protocols. Whenever self-supervised
pretraining is performed on a dataset, the listed training videos are used
without labels. These datasets are selected to enable direct comparison with
prior point-cloud-video methods, and the split protocols match the baselines
reported in the main paper. Input clips are extracted from each video with overlapping sliding windows. Training-time point sampling is
stochastic. MSRAction-3D and NTU RGB+D apply random axis-wise scaling in
$[0.9,1.1]$, while SHREC'17 applies the same scaling after clip normalization.

\begin{table}[ht]
    \centering
    \begingroup
    \setlength{\tabcolsep}{2pt}
    \begin{tabular}{@{}>{\raggedright\arraybackslash}p{0.27\columnwidth}cc>{\raggedright\arraybackslash}p{0.28\columnwidth}@{}}
        \toprule
        Dataset & Classes & Train/Test & Protocol \\
        \midrule
        MSRAction-3D & 20 & 270 / 297 &
        S1--S5 / S6--S10 \\
        NTU RGB+D & 60 & 40,320 / 16,560 &
        Official \\
        SHREC'17 & 28 & 1,960 / 840 &
        Official \\
        \bottomrule
    \end{tabular}
    \endgroup
    \caption{Dataset statistics and cross-subject split protocols used in the experiments. Counts are video-level sequence counts under the corresponding protocol.}
    \label{tab:supp-data-protocols}
\end{table}

The NTU RGB+D semi-supervised setting uses a class-balanced 50\% subset of the cross-subject training set, comprising 20,160 of the 40,320 training videos, with 336 videos per class. The MSRAction-3D few-shot setting retains all 20 action classes and samples 1, 3, or 5 labeled videos per class. In both settings, the sampled subsets are fixed throughout fine-tuning; for MSRAction-3D, each support set is also shared between the paired scratch and pretrained runs for the corresponding random seed.

\subsection{Training and Architecture Configurations}

Tables~\ref{tab:supp-training-protocols} and~\ref{tab:supp-architecture} report
the protocol-specific optimization, point-tube sampling, and architecture
configurations used for the experiments in the main paper. Across protocols, we
use AdamW with a 10-epoch linear warmup followed by cosine decay to a minimum
learning rate of $10^{-6}$, and clip gradients to a global norm of 1.0. The
training objective is Smooth L1 loss with $\beta=2.0$ for masked prediction
during self-supervised pretraining and cross-entropy loss during supervised
fine-tuning. During pretraining, we mask 75\% of point-tube tokens, use a
temporal kernel size of 3 with temporal stride 2, and apply SIGReg with 17 knots
and 1024 random projections when the regularizer is enabled.

\begin{table*}[!t]
    \centering
    \begingroup
    \small
    \setlength{\tabcolsep}{4pt}
    \begin{tabular}{@{}>{\raggedright\arraybackslash}p{0.21\textwidth}cccccccccccc@{}}
        \toprule
        \textbf{Protocol} & \textbf{Epochs} & \textbf{Batch} & \textbf{LR} & \textbf{WD} & \textbf{Frames} & \textbf{Pts.} & \textbf{Clip Str.} & \textbf{Frame Str.} & \textbf{$r_s$} & \textbf{Sp. Str.} & \textbf{Nbrs.} & \textbf{SIGReg} \\
        \midrule
        MSR pretraining & 200 & 96 & 3e-4 & 5e-2 & 24 & 1024 & 1 & 1 & 0.3 & 32 & 32 & 0.01 \\
        NTU pretraining & 100 & 56 & 3e-4 & 5e-2 & 24 & 1024 & 2 & 2 & 0.1 & 32 & 32 & 0.03 \\
        MSR full fine-tuning & 50 & 48 & 5e-4 & 1e-4 & 24 & 2048 & 1 & 1 & 0.7 & 32 & 32 & -- \\
        NTU full/semi fine-tuning & 20 & 48 & 5e-4 & 5e-2 & 24 & 2048 & 2 & 2 & 0.1 & 32 & 32 & -- \\
        SHREC transfer fine-tuning & 50 & 24 & 1e-3 & 1e-4 & 16 & 256 & 1 & 1 & 0.3 & 16 & 9 & -- \\
        MSR few-shot fine-tuning & 50 & 48 & 5e-4 & 1e-4 & 24 & 2048 & 1 & 1 & 0.7 & 32 & 32 & -- \\
        \bottomrule
    \end{tabular}
    \endgroup
    \caption{Final optimization and point-tube sampling settings. LR and WD denote learning rate and weight decay; Frames denotes frames per input clip; Pts. denotes sampled points per frame; Clip Str. and Frame Str. denote the temporal strides between successive clips and successive frames within a clip, respectively. Sp. Str. and Nbrs. denote P4D spatial stride and neighbor samples. SIGReg weights apply only to self-supervised pretraining.}
    \label{tab:supp-training-protocols}
\end{table*}

\begin{table}[ht]
    \centering
    \begingroup
    \setlength{\tabcolsep}{3pt}
    \begin{tabular}{@{}lcccc@{}}
        \toprule
        \textbf{Setting} & \textbf{MSR} & \textbf{NTU} & \textbf{SHREC} & \textbf{Few-shot} \\
        \midrule
        Enc. dim & 384 & 384 & 384 & 384 \\
        Enc. heads & 8 & 8 & 8 & 8 \\
        Enc. depth & 5 & 10 & 10 & 10 \\
        Enc. MLP & 4 & 4 & 4 & 4 \\
        Pred. dim & 192 & 192 & -- & -- \\
        Pred. heads & 6 & 6 & -- & -- \\
        Pred. depth & 3 & 6 & -- & -- \\
        Pred. MLP & 4 & 4 & -- & -- \\
        Cls. hidden & 48 & 128 & 128 & 16 \\
        Cls. dropout & 0.5 & 0.5 & 0.5 & 0.5 \\
        \bottomrule
    \end{tabular}
    \endgroup
    \caption{Architecture settings. Enc., Pred., and Cls. denote encoder, predictor, and classifier, respectively; dim, heads, depth, MLP, hidden, and dropout report the corresponding architectural hyperparameters. The NTU semi-supervised experiments use the same architecture as the full-data NTU experiments.}
    \label{tab:supp-architecture}
\end{table}

\subsection{Random Seeds}

Standard and ablation training runs use seed 42, whereas few-shot MSRAction-3D training runs use seeds 0--9. The seed setup used by each training entry point is:
\begin{center}
\begin{tabular}{@{}l@{}}
\texttt{seed = args.seed}\\
\texttt{random.seed(seed)}\\
\texttt{np.random.seed(seed)}\\
\texttt{torch.manual\_seed(seed)}\\
\texttt{torch.cuda.manual\_seed\_all(seed)}
\end{tabular}
\end{center}
This controls the pseudorandom streams used for initialization, data sampling,
augmentation, masking, and SIGReg projections; deterministic CUDA kernels are
not forced.

\subsection{Ablation Protocols}

For controlled MSRAction-3D ablations, all settings in
Tables~\ref{tab:supp-training-protocols} and~\ref{tab:supp-architecture} are
kept fixed except the studied factor: target detachment, target-position query,
mask ratio, SIGReg weight, or an EMA target encoder with momentum increasing
from 0.996 to 0.9998. The reported mask-ratio range is
$m\in\{0.65,0.75,0.85\}$ and the reported SIGReg-weight range is
$\lambda_{\mathrm{SIGReg}}\in\{0.001,0.01,0.1\}$ for the MSRAction-3D
ablations. These are targeted development ablations rather than an exhaustive
hyperparameter grid.

\subsection{Evaluation}

During evaluation, we process all clips extracted from each test video using
the sliding-window protocol. The reported video-level top-1 accuracy is computed by summing the softmax
probabilities over clips from the same video and assigning the video to the
class with the largest accumulated probability. We use this metric because all
evaluated benchmarks are single-label action or gesture classification tasks
and because video-level top-1 accuracy is the standard metric used by the
compared methods. For pretrained models, we initialize fine-tuning from the
final self-supervised pretraining checkpoint. Following the protocol used in
prior work, we compute video-level top-1 accuracy on the test split after each
fine-tuning epoch and report the best result over fine-tuning epochs.

\section{Additional Results}

\subsection{Prediction-Loss Diagnostics for Ablations}

Tables~\ref{tab:supp-pred-loss-component}
and~\ref{tab:supp-pred-loss-sensitivity} complement the corresponding component
and sensitivity ablations in the main paper. For each MSRAction-3D variant,
they report downstream video-level accuracy together with two late-pretraining
diagnostics. Prediction loss is the Smooth L1 objective between the predicted
masked latents $\hat{\mathbf{Y}}$ and the layer-normalized targets $\mathbf{Y}$,
as defined in the main paper. Target Std.\ is included as a collapse diagnostic
and denotes the mean per-feature standard deviation of $\mathbf{Y}$ after
flattening its non-feature dimensions. Both diagnostics are averaged over
minibatches and the final ten pretraining epochs.

\begin{table}[ht]
    \centering
    \begingroup
    \small
    \setlength{\tabcolsep}{2pt}
    \begin{tabular}{@{}>{\raggedright\arraybackslash}p{0.46\columnwidth}rrr@{}}
        \toprule
        \textbf{Variant} & \textbf{Pred. Loss} & \textbf{Target Std.} & \textbf{Acc.} \\
        \midrule
        Full model ($m=0.75$, $\lambda_{\mathrm{SIGReg}}=0.01$) & 0.001219 & 0.984 & \textbf{94.08} \\
        w/o SIGReg & 0.000000 & 0.008 & 89.90 \\
        Stop-gradient targets & 0.095069 & 0.993 & 79.44 \\
        w/o target-position query & 0.002563 & 0.942 & 93.03 \\
        EMA target encoder & 0.000452 & 0.388 & 88.50 \\
        \bottomrule
    \end{tabular}
    \endgroup
    \caption{Prediction-loss diagnostics for component/path ablations on MSRAction-3D.}
    \label{tab:supp-pred-loss-component}
\end{table}

\begin{table}[ht]
    \centering
    \begingroup
    \small
    \setlength{\tabcolsep}{2pt}
    \begin{tabular}{@{}>{\raggedright\arraybackslash}p{0.27\columnwidth}crrr@{}}
        \toprule
        \textbf{Setting} & \textbf{Value} & \textbf{Pred. Loss} & \textbf{Target Std.} & \textbf{Acc.} \\
        \midrule
        Mask ratio $m$ & 0.65 & 0.001050 & 0.984 & 88.85 \\
        Mask ratio $m$ & 0.75 & 0.001219 & 0.984 & \textbf{94.08} \\
        Mask ratio $m$ & 0.85 & 0.001401 & 0.985 & 91.99 \\
        \midrule
        $\lambda_{\mathrm{SIGReg}}$ & 0.001 & 0.000179 & 0.870 & 86.76 \\
        $\lambda_{\mathrm{SIGReg}}$ & 0.01 & 0.001219 & 0.984 & \textbf{94.08} \\
        $\lambda_{\mathrm{SIGReg}}$ & 0.1 & 0.010965 & 0.992 & 89.55 \\
        \bottomrule
    \end{tabular}
    \endgroup
    \caption{Prediction-loss diagnostics for sensitivity ablations on MSRAction-3D.}
    \label{tab:supp-pred-loss-sensitivity}
\end{table}

These diagnostics indicate that prediction loss alone is not a reliable model
selection criterion. Without SIGReg, the model obtains a near-zero logged
prediction loss, but the target-latent standard deviation collapses and
downstream accuracy decreases. The EMA target-encoder variant also has a small
prediction loss, but its target variance remains substantially lower than that
of the shared-encoder SIGReg formulation. In contrast, stop-gradient targets
retain near-unit target variance, but make masked-target prediction harder and
produce the lowest downstream accuracy. The full model therefore performs best
not because it minimizes prediction loss, but because it maintains a
non-collapsed latent space while preserving an informative predictive objective.

The sensitivity ablations reveal distinct trade-offs. A low mask ratio provides
a weaker prediction task, whereas a high ratio removes too much visible context;
moderate masking therefore performs best, while target variance remains stable
across settings. For SIGReg, weak regularization is insufficient to maintain
target variance, whereas strong regularization raises prediction loss, indicating
a trade-off with predictive alignment. An intermediate weight provides the
best downstream balance.

\begin{table*}[t]
    \centering
    \begingroup
    \small
    \setlength{\tabcolsep}{4pt}
    \begin{tabular}{@{}lrrrrrr@{}}
        \toprule
        & \multicolumn{2}{c}{\textbf{1-shot}} & \multicolumn{2}{c}{\textbf{3-shot}} & \multicolumn{2}{c}{\textbf{5-shot}} \\
        \cmidrule(lr){2-3}\cmidrule(lr){4-5}\cmidrule(l){6-7}
        \textbf{Seed} & \textbf{Scratch} & \textbf{Pretrained} & \textbf{Scratch} & \textbf{Pretrained} & \textbf{Scratch} & \textbf{Pretrained} \\
        \midrule
        0 & 35.889 & 33.798 & 62.718 & 62.718 & 79.094 & 77.003 \\
        1 & 19.861 & 27.526 & 24.739 & 65.505 & 77.700 & 72.474 \\
        2 & 30.662 & 38.328 & 57.840 & 60.279 & 84.669 & 83.275 \\
        3 & 6.969 & 36.585 & 54.355 & 58.537 & 80.139 & 81.882 \\
        4 & 13.589 & 37.631 & 40.767 & 64.460 & 82.927 & 81.185 \\
        5 & 5.923 & 13.589 & 43.206 & 57.143 & 66.899 & 74.913 \\
        6 & 21.603 & 29.268 & 79.094 & 75.261 & 79.094 & 86.063 \\
        7 & 31.707 & 30.314 & 74.564 & 67.596 & 77.352 & 80.836 \\
        8 & 12.544 & 43.902 & 5.226 & 64.460 & 49.826 & 81.533 \\
        9 & 21.951 & 19.512 & 55.401 & 58.188 & 81.185 & 78.397 \\
        Mean $\pm$ Std. & $20.070{\pm}10.404$ & $31.045{\pm}9.152$ & $49.791{\pm}22.348$ & $63.415{\pm}5.428$ & $75.889{\pm}10.321$ & $79.756{\pm}4.068$ \\
        \bottomrule
    \end{tabular}
    \endgroup
    \caption{Per-seed all-class few-shot P4Transformer video-level top-1 accuracy (\%) on MSRAction-3D. Means are reported with standard deviations over ten run seeds.}
    \label{tab:supp-few-shot-multirun}
\end{table*}

\subsection{Few-Shot Multi-Run Results}

Table~\ref{tab:supp-few-shot-multirun} reports the per-run MSRAction-3D few-shot
results used to summarize robustness across support-set samples. Scratch denotes
training without a pretrained checkpoint, whereas Pretrained uses the
self-supervised checkpoint for initialization. Each seed uses the same labeled
support set for both initializations, so the two columns form paired
comparisons.

\end{document}